\RequirePackage{fix-cm}
\documentclass[smallextended]{svjour3}       
\smartqed  
\usepackage{graphicx}

\usepackage{graphicx}
\usepackage{amsmath}
\usepackage{amssymb}

\usepackage{subfigure}
\usepackage{array}
\usepackage{multirow,tabularx}
\usepackage{multicol}

\newcommand{\bfx}{{\textbf{x}}}
\newcommand{\bfv}{{\textbf{v}}}

\newcommand{\bfw}{{\textbf{w}}}
\newcommand{\bfy}{{\textbf{y}}}

\newcommand{\bfz}{{\textbf{z}}}

\newcommand{\bfalpha}{{\boldsymbol{\alpha}}}

\newcommand{\bfeta}{{\boldsymbol{\eta}}}

\begin{document}

\title{Cross-model convolutional neural network for multiple modality data representation}

\author{Yanbin Wu \and Li Wang \and Fan Cui \and Hongbin Zhai \and Baoming Dong \and Jim Jing-Yan Wang
}

\institute{Yanbin Wu, Hongbin Zhai \at
College of Management Science \& Engineering, Hebei University of Economics and Business, Shijiazhuang 050061, China
\and
Li Wang\at
State Key Laboratory of Remote Sensing Science, Institute of Remote Sensing and Digital Earth, Chinese Academy of Sciences, Beijing 100101, China
\and
Fan Cui\at
University Party and Government Office, China University of Mining and Technology (Beijing), Beijing 100083, China
\and
Baoming Dong\at
College of Business Administration, Hebei University of Economics and Business, Shijiazhuang 050061, China\\
Baoming Dong is the corresponding author.\\
\email{baomingdong1@outlook.com}
\and
Jim Jing-Yan Wang\at
New York University Abu Dhabi, Abu Dhabi, United Arab Emirates
}

\date{Received: date / Accepted: date}

\maketitle

\begin{abstract}
A novel data representation method of convolutional neural network (CNN) is proposed in this paper to represent data of different modalities. We learn a CNN model for the data of each modality to map the data of different modalities to a common space, and regularize the new representations in the common space by a cross-model relevance matrix. We further impose that the class label of data points can also be predicted from the CNN representations in the common space. The learning problem is modeled as a minimization problem, which is solved by an augmented Lagrange method (ALM) with updating rules of Alternating direction method of multipliers (ADMM). The experiments over benchmark of sequence data of multiple modalities show its advantage.
\end{abstract}

\section{Introduction}

\subsection{Background}

In the machine learning community, the convolutional neural network (CNN) has been a popular method to represent data of sequence \cite{krizhevsky2012imagenet,lawrence1997face,simard2003best,kalchbrenner2014convolutional,ciresan2011convolutional,qin2010progressive,yuan2016deep,li2016nuclear}. Given a sequence of instances, CNN uses a sliding window to split the sequence into a group of short sub-sequences. Then it uses a bank of filters to filter the instances in these sub-sequences, and finally performs a max-pooling to the outputs of different sub-sequences. The outputs corresponding to the filters in the filter bank are used as the new representations of the sequence for the problem of classification and retrieval. Recently, CNN model has been extensively explored to represent different types of sequence data, such as image, text, video, and protein. For example, in the natural language problems, each sentence is treated as a sequence of words, and each word is represented by a word embedding vector. The sequence of word embedding vectors can be represented further by a CNN model for the problem of sematic classification \cite{fan2016stochastic,dos2014deep,collobert2008unified}. Moreover, in computer vision applications, a video is also composed of a sequence of image frames, and we can also extract visual feature vector from each frame. In this case, we can also use d CNN model to represent the sequence of video for the problem of scenes classification or objective classification \cite{jia2014caffe,ciregan2012multi,ciresan2011flexible,liang2016novel}.

Meanwhile, with the rapid development of internet technology, the social network is becoming more and more popular. In the social network, a great amount of data is being generated. The modalities of the data is usually of different types. For example, on a webpage of a Facebook profile, there are text, image, video, etc \cite{wang2015supervised}. The diversity of the modalities of the data make the problem of classification and retrieval even more complex and difficult. To process and understand the multiple modality data, it is necessary to develop cross-model data representation, classification, and retrieval methods. However, up to now, all the existing CNN methods are limited to single modality data. When one CNN is applied to one modality data, other modality data are ignored. It is possible to learn independent CNN models for different modalities, but the sematic similarity/relevance of data points of different modalities are ignored. However, in current internet data analysis, retrieval, and understanding applications, the cross modality sematic relevance plays critical role. For example, in the multimedia retrieval application, the users usually use a textual description as query to retrieve image and/or video data. In this application, it is necessary to explore cross modality information.

\subsection{Relevant works}

To handle the multiple modality data, many cross-model data representation methods have been proposed. For example,

\begin{itemize}
\item Wang et al. \cite{He7346492} proposed a cross-model joint feature selection and subspace learning method to represent data of different modalities. The matrices of different modalities are projected to a common subspace by subspace learning, and in this common subspace the similarity between data points of different modalities can be measured. Moreover, the projection matrices are also regularized by the $\ell_2$ norm penalties to select the relevant and discriminative features. To bridge the data points of different modalities, a multi-modal graph regularization term is used to regularize the data in the common space, and it can preserves the inter-modality and intra-modality similarity. The joint learning problem is solved by an iterative algorithm.
\item Masci et al. \cite{masci2014multimodal} proposed a hashing method to represent data of different modalities and used it to map the data points of these modalities to a common space, so that they can be compared in this common space. The proposed method also use the intra- and inter-modality similarity to regularize the learning of the hashing parameters.
\item Li et al. \cite{Li7571151} proposed a ranking-based hashing to map the data of different modalities to a common space so that the similarity between different modalities can be measured by Hamming distance. The hashing function is not a traditional sign or threshold function, by a max-pooling function based on rank correlation measures. This hashing function is a natural probabilistic approximation. The hashing parameters are also learned by the regularization of the intra- and inter-similarity of different modalities.
\end{itemize}

\subsection{Our contributions}

Although there are some works on cross-model representation, but they are all limited to the subspace learning methods. For example, the method of Wang et al. \cite{He7346492} is a subspace learning method, and the method of Li et al. \cite{Li7571151} is a subspace learning-based hashing method. It is still unknown how the CNN model performs for the problem of cross-model representation problem. To fill this gap, in this paper, we propose the first cross-model CNN learning method. We learn a CNN model for each modality, and then use the CNN models to map the data points of different modalities to a common space. In this common space, we further propose to learn a linear classifier to predict the class label from the CNN representations. Moreover, we also propose to use the cross-modal relevance matrix to regularize the learning of the CNN representations. If two data points of either the same modality or different modalities are relevant in the sematic concept, their CNN representations are imposed to be close to each other, and vice versa. The parameters of the classifier and the CNN models are also regularized by $\ell_2$ norms to reduce the complexity. The objective function is a joint framework of these considered problem, and we develop an algorithm of augmented Lagrangian method (ALM) to minimize the objective function. The experiments over multi-modality data sets for retrieval problem show the proposed method outperforms better than existing cross-modal data representation methods.

The rest parts of this paper is organized as follows. In section \ref{sec:proposed}, we introduce the proposed method of cross model CNN. In section \ref{sec:exp}, the proposed method is evaluated. In section \ref{sec:conc}, the conclusions of this paper are given.

\section{Proposed method}
\label{sec:proposed}

\subsection{Problem modeling}

In many multiple modality data processing applications, each data point is presented as a sequence of instances, $\mathcal{X}=\{\bfx_1,\cdots,\bfx_{|\mathcal{X}|}\}$, where $\bfx_\tau \in \mathbb{R}^d$ is the $d$-dimensional input vector of the $\tau$-th instances. For example, for the modality of text, each text is given as a sequence of words, and each work can be presented as a embedding vector. For the modality of video, each video is a sequence of frames, and we can extract a visual feature vector for each frame. To represent the data point, we first use a sliding window to explore the context information for each instance. The size of the sliding window is denoted as $h$, and the window covers the neighboring $h$ instances. The window slides from the beginning of $\mathcal{X}$ to its end, and generates a sequence of sub-sequences of instances,

\begin{equation}
\begin{aligned}
(\bfx_1,\cdots,\bfx_{h}), \cdots,(\bfx_\tau,\cdots,\bfx_{\tau+h-1}), \cdots,(\bfx_{|\mathcal{X}|-h+1},\cdots,\bfx_{|\mathcal{X}|}).
\end{aligned}
\end{equation}
In the $\tau$-th sequence, the $\tau$-th instance to the $\tau+h-1$ is included as $(\bfx_\tau,\cdots,\bfx_{\tau+h-1})$. In this way, the contextual instances of each instance in the sequence are explored and used as its new presentation. We further concatenate them to a $d\times h$-dimensional longer vector,

\begin{equation}
\begin{aligned}
\bfy_\tau = [\bfx_\tau^\top,\cdots,\bfx_{\tau+h-1}^\top]^\top \in \mathbb{R}^{dh}.
\end{aligned}
\end{equation}
In this way, $\mathcal{X}$ is represented as a sequences of $|\mathcal{X}|-h+1$ windows,

\begin{equation}
\begin{aligned}
\mathcal{Y} =\{ \bfy_1,\cdots,\bfy_{|\mathcal{X}|-h+1}\}.
\end{aligned}
\end{equation}
Then we perform convolution operation to this sequence of window vectors with a filter $\bfw\in \mathbb{R}^{dh}$. The convolution operation is composed of a sequence of pairs of filtering and nonlinear transformation actions. For a window representation, $\bfy_\tau$, the output of the filtering action is the dot-product between $\bfw$ and $\bfy_\tau$, $\bfw^\top\bfy_\tau$, and the online transformation function is denoted as $\sigma(\cdot)$, and it is defined as a tanh function in our paper, $\sigma(x) = tanh(x)$. Thus the output of the filtering-nonlinear transformation action is $\sigma(\bfw^\top\bfy_\tau)$, and the outputs of the convolution is given as follows,

\begin{equation}
\begin{aligned}
\{\sigma(\bfw^\top\bfy_1),\cdots,\sigma(\bfw^\top\bfy_{|\mathcal{X}|-h+1})\}.
\end{aligned}
\end{equation}
Then the max-pooling operation is performed to the sequence of the outputs of the convolution operation to select the maximum output,
\begin{equation}
\label{equ:z}
\begin{aligned}
z = \max_{\tau=1}^{|\mathcal{X}|-h+1}\sigma(\bfw^\top\bfy_\tau).
\end{aligned}
\end{equation}
Actually, we will use a filter bank of multiple filters for the convolution operation, denoted as $\mathcal{W}=\{\bfw_1,\cdots,\bfw_{u}\}$, where $u$ is the number of filters in the filter bank. The outputs generated by the $k$-th filter is denoted as $z_k$, $z_k = \max_{\tau=1}^{|\mathcal{X}|-h+1}\sigma(\bfw_k^\top\bfy_\tau)$, and the outputs with regard to the filters in $\mathcal{W}$ are concatenated to form a vector of the convolutional outputs,

\begin{equation}
\begin{aligned}
\bfz = [z_1,\cdots,z_{u}]^\top\in \mathbb{R}^{u}.
\end{aligned}
\end{equation}
This vector is a new representation of the data point $\mathcal{X}$. Moreover, to predict its binary label $\eta\in \{+1,-1\}$, we further apply a linear classification function $f(\cdot)$ to this representation,

\begin{equation}
\begin{aligned}
\eta\leftarrow f(\bfz;\bfv)=\bfv^\top\bfz=\sum_{k=1}^{u} v_k z_k,
\end{aligned}
\end{equation}
where $\bfv=[z_1,\cdots,z_{u}]\in R^{u}$ is the parameter of the classification function.

Please note that for different modalities, we use different filter banks, but the same classifier. In this way, the sequences of different modalities are mapped to a common CNN space, and in this space, a common classifier can be applied. To conduct the cross-model retrieval and classification, we propose to use the CNN model to map the data points of different modalities to a common CNN representation space. The filter bank for the $j$-th modality is denoted as $\mathcal{W}_j = \{\bfw_1^j,\cdots,\bfw_{u}^j\}$, where $\bfw_k^j$ is the $k$-th filter of $\mathcal{W}_j$. Please note that the sizes of the filter bank of different modalities are the same, $u$. The classifier parameter is denoted as $\bfv$. The learning algorithm is to learn both the filter bank, $\mathcal{W}_j$, for each modality and the cross-model classifier parameter, $\bfv$. To this end, we use a training data set to learn the parameters. We assume we have a training set of $m$ modalities, and the training subset of the $j$-th modality is denoted as $\mathcal{D}_j,j=1,\cdots,m$. $\mathcal{D}_j = \{\mathcal{X}^j_1,\cdots,\mathcal{X}^j_{|\mathcal{D}_j|}\}$ is composed of $|\mathcal{D}_j|$ data points, and $\mathcal{X}^j_i$ is its $i$-th data point. We use the proposed CNN model to represent the data point $\mathcal{X}_i^j$ to a vector $\bfz_i^j$. The following problems are considered in the training process.

\begin{itemize}
\item To approximate the class labels from the CNN representations correctly for the training data points, we propose to minimize the loss function for all the modalities. The loss function for the classification over the $i$-th data point of the $j$-th modality, $\mathcal{X}_i^j$, is given as the squared loss as follows,

\begin{equation}
\begin{aligned}
\ell(\eta_i^j,\bfv^\top\bfz_i^j) = \|\eta_i^j - \bfv^\top\bfz_i^j\|_2^2,
\end{aligned}
\end{equation}
where $\eta_i^j\in \{+1,-1\}$ is the ground truth binary label of  $\mathcal{X}_i^j$. To approximate the ground truth label as correctly as possible, we propose to minimize the loss function over all the training data points of different modalities,

\begin{equation}
\begin{aligned}
\min\sum_{j=1}^m\sum_{i=1}^{|\mathcal{D}_j|} \ell(\eta_i^j,\bfv^\top\bfz_i^j).
\end{aligned}
\end{equation}

\item To prevent the problem of over-fitting to the training data, we propose to regularize the parameters of classifiers and the filters. The regularization term is designed as their squared $\ell_2$ norms. The regularization terms are also minimized to seek a solution as simple as possible,

\begin{equation}
\begin{aligned}
\min\left (\|\bfv\|_2^2 + \sum_{j=1}^m\sum_{k=1}^{u} \|\bfw_k^j\|_2^2 \right ).
\end{aligned}
\end{equation}

\item To bridge the information of different modalities, we propose that the CNN representations of data points of different modalities should share the same data space, and we use a sematic relevance matrix to regularize the representations in this space. The relevance between two data points $\mathcal{X}_i^j$ of the $j$-th modality, and $\mathcal{X}_{i'}^{j'}$ of the $j'$-th modality is given as a binary value, $S_{ii'}^{jj'}\in \{1,0\}$,

\begin{equation}
\begin{aligned}
S_{ii'}^{jj'} =
\left\{\begin{matrix}
1, &if~\mathcal{X}_i^j~and~\mathcal{X}_{i'}^{j'}~are~semantically~relevant,\\
-1, &if~\mathcal{X}_i^j~and~\mathcal{X}_{i'}^{j'}~are~semantically~irrelevant,~and\\
0, &if~their~relevance~is~unknown.
\end{matrix}\right.
\end{aligned}
\end{equation}
This sematic similarity describes the relevance between two data points from either the same modality or different modalities. Naturally, we hope the dissimilarity between the two CNN representations of $\mathcal{X}_i^j$ and $\mathcal{X}_{i'}^{j'}$, $\bfz_i^j$ and $\bfz_{i'}^{j'}$, are as small as possible if $S_{ii'}^{jj'}=1$. For the $\mathcal{X}_i^j$ and $\mathcal{X}_{i'}^{j'}$ pair with $S_{ii'}^{jj'}=-1$, we hope their dissimilarity can be as big as possible. The dissimilarity between $\bfz_i^j$ and $\bfz_{i'}^{j'}$ is given as the squared $\ell_2$ norm distance between them, $\|\bfz_i^j - \bfz_{i'}^{j'}\|_2^2$, and the following minimization problem is argued,
\begin{equation}
\begin{aligned}
\min \sum_{j,j'=1}^m \left ( \sum_{i}^{|\mathcal{D}_j|} \sum_{i'=1}^{|\mathcal{D}_j|}
S_{ii'}^{jj'} \|\bfz_i^j - \bfz_{i'}^{j'}\|_2^2 \right ).
\end{aligned}
\end{equation}
This minimization is a cross-model matching regularization term. It not only requires that the data points from the same modality should be regularized by the sematic relevance, but also apply the sematic regularization to the data from different modalities.

\end{itemize}

Summarizing the problems above, we obtain the optimization problem for the cross-model CNN learning,

\begin{equation}
\label{equ:objective}
\begin{aligned}
&\min_{\bfv,\mathcal{W}_{j,j=1,\cdots,m}}
\left\{\sum_{j=1}^m\sum_{i=1}^{|\mathcal{D}_j|} \ell(\eta_i^j,\bfv^\top\bfz_i^j)
+\lambda_1 \left (\|\bfv\|_2^2 + \sum_{j=1}^m\sum_{k=1}^{u} \|\bfw_k^j\|_2^2 \right )
\right.\\
&
\left.
+\lambda_2 \sum_{j,j'=1}^m \left ( \sum_{i}^{|\mathcal{D}_j|} \sum_{i'=1}^{|\mathcal{D}_{j'}|}
S_{ii'}^{jj'} \|\bfz_i^j - \bfz_{i'}^{j'}\|_2^2 \right )
\right \},
\end{aligned}
\end{equation}
where $\lambda_1$ and $\lambda_2$ are tradeoff parameters of the last two regularization terms. They control the influences of the regularization of these two terms over the final solutions. Their values are decided by linear search in our experiments. Please note that the CNN representations are functions of the filter banks according to (\ref{equ:z}). In (\ref{equ:z}), a nonlinear function and a max-pooling operation is coupled. Direct optimization of the filters by solving (\ref{equ:objective}) is difficult. Instead of solving the filters directly, we explicitly introduce the CNN representation vectors to the optimization problem as independent variables, and put a constraint to impose the relation between the CNN representations and the CNN functions, $[\bfz_i^j]_k = \max_{\tau=1}^{|\mathcal{X}_i^j|-h+1}\sigma({\bfw_k^j}^\top[\bfy_i^j]_\tau)$, where $[\bfz_i^j]_k$ is the $k$-th element of the vector $\bfz_i^j$. The reformed optimization problem is given as follows,

\begin{equation}
\label{equ:objective1}
\begin{aligned}
&\min_{\bfv,\left(\mathcal{W}_j,\bfz^j_{i,i=1,\cdots,|\mathcal{D}_j}|\right)_{j=1,\cdots,m}}
\left\{\sum_{j=1}^m\sum_{i=1}^{|\mathcal{D}_j|} \ell(\eta_i^j,\bfv^\top\bfz_i^j)
\right.\\
&
\left.
+\lambda_1 \sum_{j=1}^m \left (\|\bfv\|_2^2 + \sum_{k=1}^{u} \|\bfw_k^j\|_2^2 \right )
+\lambda_2 \sum_{j,j'=1}^m \left ( \sum_{i}^{|\mathcal{D}_j|} \sum_{i'=1}^{|\mathcal{D}_{j'}|}
S_{ii'}^{jj'} \|\bfz_i^j - \bfz_{i'}^{j'}\|_2^2 \right )
\right \},\\
&subject~to~[\bfz_i^j]_k = \max_{\tau=1}^{|\mathcal{X}_i^j|-h+1}\sigma({\bfw_k^j}^\top[\bfy_i^j]_\tau),\forall~i,j,k.
\end{aligned}
\end{equation}
This optimization problem is a constrained minimization problem. It is interesting to note that in the objective function of this problem, we only have variables of the classifier parameter, the CNN filters, and the CNN representation vectors, the labels of the training data, and the cross-model similarity matrix as the input. The input sequences are not included in the objective function, but only appear in the constraints.

\subsection{Problem optimization}

To solve the problem in (\ref{equ:objective1}), we use the ALM method \cite{yang2011alternating}. The augmented Lagrangian function of the problem of (\ref{equ:objective1}) is given as follows,

\begin{equation}
\label{equ:Lagrangian}
\begin{aligned}
&\mathcal{L}(\mathcal{W}_j,\bfv,\bfz_i^j; \alpha_{ik}^k)=
\sum_{j=1}^m\sum_{i=1}^{|\mathcal{D}_j|} \ell(\eta_i^j,\bfv^\top\bfz_i^j)
\\
&
+\lambda_1 \left (\|\bfv\|_2^2 + \sum_{j=1}^m \sum_{k=1}^{u} \|\bfw_k^j\|_2^2 \right )
+\lambda_2 \sum_{j,j'=1}^m \left ( \sum_{i}^{|\mathcal{D}_j|} \sum_{i'=1}^{|\mathcal{D}_{j'}|}
S_{ii'}^{jj'} \|\bfz_i^j - \bfz_{i'}^{j'}\|_2^2 \right )
\\
&+\sum_{j=1}^m\sum_{i=1}^{|\mathcal{D}_j|}\sum_{k=1}^{u}\alpha_{ik}^j\left([\bfz_i^j]_k - \max_{\tau=1}^{|\mathcal{X}_i^j|-h+1}\sigma({\bfw_k^j}^\top[\bfy_i^j]_\tau)\right)\\
&+\frac{\beta}{2}\sum_{j=1}^m\sum_{i=1}^{|\mathcal{D}_j|}\sum_{k=1}^{u}\left([\bfz_i^j]_k - \max_{\tau=1}^{|\mathcal{X}_i^j|-h+1}\sigma({\bfw_k^j}^\top[\bfy_i^j]_\tau)\right)^2,
\end{aligned}
\end{equation}
where $\alpha_{ik}^j$ is the Lagrange multiplier of the constraint $[\bfz_i^j]_k = \max_{\tau=1}^{|\mathcal{X}_i^j|-h+1}\sigma({\bfw_k^j}^\top[\bfy_i^j]_\tau)$, and $\beta$ is its positive penalty parameter. Alternating direction method of multipliers (ADMM) \cite{boyd2011distributed} is used to solve the problem respect to the filters in $\mathcal{W}_j$, the classifier parameters, $\bfv$, the CNN representations, $\bfz_i^j$, and the Lagrange multipliers, $\alpha_{ik}^j$ jointly. The ADMM algorithm update these variables sequentially, and the updating steps are given in Algorithm 1.

\begin{itemize}
\item \textbf{Algorithm 1}: ADMM algorithm to update the variables of CNN model and the Lagrange multipliers.
\item Inputs: The sequence of instances of $m$ modalities and the corresponding labels, $\mathcal{D}_j = \{(\mathcal{X}^j_1,\eta^j_1),\cdots,(\mathcal{X}^j_{|\mathcal{D}_j|},\eta^j_{|\mathcal{D}_j|})\}$, $j=1,\cdots,m$. The sematic similarity matrix of cross-modalities, $S$. The tradeoff parameters $\lambda_1$ and $\lambda_2$.
\item Initialize $\bfv^0$, $\mathcal{W}_j^0$, $(\bfz_i^j)^0$, and $(\alpha_{ik}^j)^0$, for $j=1,\cdots,m$, $i=1,\cdots,|\mathcal{D}_j|$, and $k=1,\cdots,u$.
\item While not converged do
\begin{enumerate}
  \item $\bfv^{t+1} = {\arg\min}_{\bfv} \mathcal{L}(\mathcal{W}_j^t,\bfv,(\bfz_i^j)^t;(\alpha_{ik}^k)^t)$, $j=1,\cdots,m$.
  \item $(\bfz_i^j)^{t+1} = {\arg\min}_{\bfz^j_i} \mathcal{L}(\mathcal{W}_j^t,\bfv^{t+1},\bfz_i^j;(\alpha_{ik}^k)^t)$, $j=1,\cdots,m$, $i=1,\cdots,|\mathcal{D}_j|$.
  \item $\mathcal{W}_j^{t+1} = {\arg\min}_{\mathcal{W}_j} \mathcal{L}(\mathcal{W}_j,\bfv^{t+1},(\bfz_i^j)^{t+1};(\alpha_{ik}^k)^t)$, $j=1,\cdots,m$.
  \item $(\alpha_{ik}^j)^{t+1} = (\alpha_{ik}^j)^{t} + \beta\left([\bfz_i^j]_k^{t+1} - \max_{\tau=1}^{|\mathcal{X}_i^j|-h+1}\sigma((\bfw_k^{t+1})^\top[\bfy_i^j]_\tau)\right)$
\end{enumerate}
\end{itemize}

\subsubsection{Update Step for $\bfv$}

The first sub-optimization problem is the minimization of $\mathcal{L}(\mathcal{W}_j^t,\bfv,(\bfz_i^j)^t;(\alpha_{ik}^k)^t)$ with respect to $\bfv$. It has a quadratic form with regard to $\bfv$. The Lagrange function is reduced to the following function by removing the terms irrelevant to $\bfv$ as follows,

\begin{equation}
\label{equ:derivative1}
\begin{aligned}
h(\bfv)
&= \sum_{j=1}^{m}\sum_{i=1}^{|\mathcal{D}_j|} \ell(\eta_i^j,\bfv^\top\bfz_i^j)
+\lambda_1 \|\bfv\|_2^2\\
&= \sum_{j=1}^{m}\sum_{i=1}^{|\mathcal{D}_j|} \|\eta_i^j-\bfv^\top\bfz_i^j\|_2^2
+\lambda_1 \|\bfv\|_2^2.
\end{aligned}
\end{equation}
We solve it by setting the derivative of $h(\bfv)$ with regard to $\bfv$ to zero,

\begin{equation}
\label{equ:derivative2}
\begin{aligned}
&\frac{\partial h(\bfv)}{\partial \bfv}
= 2\sum_{j=1}^{m}\sum_{i=1}^{|\mathcal{D}_j|} \bfz_i^j ({\bfz_i^j}^\top \bfv- \eta_i^j)
+2\lambda_1 \bfv=0,\\
&\bfv^* = \left(\sum_{j=1}^{m}\sum_{i=1}^{|\mathcal{D}_j|} \bfz_i^j {\bfz_i^j}^\top + \lambda_1  I\right)^{-1}\left(\sum_{j=1}^{m}\sum_{i=1}^{|\mathcal{D}_j|}\bfz_i^j\eta_i^j\right).
\end{aligned}
\end{equation}

\subsubsection{Update Step for $\bfz_i^j$}

The second sub-optimization problem is the minimization of the Lagrange function with regard to $\bfz_i^j$. We remove the irrelevant terms with regard to $\bfz_i^j$ and have the objective function as follows,

\begin{equation}
\label{equ:derivative3}
\begin{aligned}
g(\bfz_i^j)
&= \sum_{j=1}^m\sum_{i=1}^{|\mathcal{D}_j|}\ell(\eta_i^j,\bfv^\top\bfz_i^j)+
\lambda_2 \sum_{j=1}^m  \sum_{j'=1}^m \left ( \sum_{i}^{|\mathcal{D}_j|} \sum_{i'=1}^{|\mathcal{D}_{j'}|}
S_{ii'}^{jj'} \|\bfz_i^j - \bfz_{i'}^{j'}\|_2^2\right)\\
&+\sum_{j=1}^m\sum_{i=1}^{|\mathcal{D}_j|}\sum_{k=1}^{u}\alpha_{ik}^j\left([\bfz_i^j]_k - \max_{\tau=1}^{|\mathcal{X}_i^j|-h+1}\sigma({\bfw_k^j}^\top[\bfy_i^j]_\tau)\right)\\
&+\frac{\beta}{2}\sum_{j=1}^m\sum_{i=1}^{|\mathcal{D}_j|}\sum_{k=1}^{u}\left([\bfz_i^j]_k - \max_{\tau=1}^{|\mathcal{X}_i^j|-h+1}\sigma({\bfw_k^j}^\top[\bfy_i^j]_\tau)\right)^2\\
&= \sum_{j=1}^m\sum_{i=1}^{|\mathcal{D}_j|}\left\|\eta_i^j-\bfv^\top\bfz_i^j\right\|^2_2+
\lambda_2 \sum_{j=1}^m  \sum_{j'=1}^m \left ( \sum_{i}^{|\mathcal{D}_j|} \sum_{i'=1}^{|\mathcal{D}_{j'}|} S_{ii'}^{jj'} \left\|\bfz_i^j - \bfz_{i'}^{j'}\right\|_2^2\right)\\
&+\sum_{j=1}^m\sum_{i=1}^{|\mathcal{D}_j|}{\bfalpha_{i}^j}^\top\left(\bfz_i^j - \overline{\bfz}_i^j\right)
+\frac{\beta}{2}\sum_{j=1}^m\sum_{i=1}^{|\mathcal{D}_j|}\left\|\bfz_i^j - \overline{\bfz}_i^j\right\|_2^2.
\end{aligned}
\end{equation}
where $\bfalpha_{i}^j = [\alpha_{i1}^j,\cdots,\alpha_{iu}^j]^\top\in \mathbb{R}^u$, and

\begin{equation}
\label{equ:zbar}
\begin{aligned}
\overline{\bfz}_i^j =\left [\max_{\tau=1}^{|\mathcal{X}_i^j|-h+1}\sigma(\bfw_1^\top[\bfy_i^j]_\tau), \cdots, \max_{\tau=1}^{|\mathcal{X}_i^j|-h+1}\sigma(\bfw_u^\top[\bfy_i^j]_\tau)\right]^\top\in \mathbb{R}^u.
\end{aligned}
\end{equation}
We defined a matrix by combining all the CNN representation vectors as the columns,

\begin{equation}
\label{equ:Z}
\begin{aligned}
Z=\left[\bfz_1^1,\cdots,\bfz^1_{|\mathcal{D}_1|}, \cdots, \bfz^{m}_1,\cdots,\bfz_{|\mathcal{D}_m|}^m \right]\in \mathbb{R}^{u\times \theta}
\end{aligned}
\end{equation}
where its $(\sum_{j'=1}^{j-1}|D_{j'}|+i)$-th column is the CNN representation vector of the $i$-th data point of the $j$-th modality, $\bfz_i^j$, and $\theta=\sum_{j=1}^{|\mathcal{D}_j|}$ is the total number of data points from all the data sets of different modalities. Similarly, we also define a label vector, a baseline CNN representation matrix, and a vector of Lagrange multipliers,

\begin{equation}
\begin{aligned}
&\overline{Z}=\left[\overline{\bfz}_1^1,\cdots,\overline{\bfz}^1_{|\mathcal{D}_1|}, \cdots, \overline{\bfz}^{m}_1,\cdots,\overline{\bfz}_{|\mathcal{D}_m|}^m \right]\in \mathbb{R}^{u\times \theta}\\
&\bfeta = [\eta_1^1,\cdots,\eta_{|\mathcal{D}_1|}^1,\cdots,\eta^m_1,\cdots,
\eta^m_{|\mathcal{D}_m|}]\in\{+1,-1\}^{\theta},~and\\
&A = [\bfalpha^1_1,\cdots,\bfalpha^1_{|\mathcal{D}_1|},\cdots,\bfalpha^m_1,\cdots,\bfalpha^m_{|\mathcal{D}_m|}]\in R^{u\times \theta}.
\end{aligned}
\end{equation}
We further define a matrix of $\theta\times \theta$ of relevance matrix for all the data points of different modalities,

\begin{equation}
\begin{aligned}
S=
\begin{bmatrix}
S_{11}^{11} & \cdots & S_{1|\mathcal{D}_1|}^{11} &   & S^{1m}_{11}  & \cdots &S^{1m}_{1|\mathcal{D}_m|}  \\
\vdots   &\ddots  & \vdots   & \cdots & \vdots   & \ddots &\vdots  \\
S_{|\mathcal{D}_1|1}^{11} & \cdots & S^{11}_{|\mathcal{D}_1||\mathcal{D}_1|} &  & S^{1m}_{|\mathcal{D}_1|1}  & \cdots &S^{1m}_{|\mathcal{D}_1||\mathcal{D}_m|}  \\
 & \vdots &  & \ddots &  & \vdots &  \\
S^{m1}_{11}  & \cdots &S^{m1}_{1|\mathcal{D}_1|}  &   &S^{mm}_{11}  & \cdots & S^{mm}_{1|\mathcal{D}_m|}\\
\vdots  & \ddots  & \vdots  & \cdots &\vdots   &  \ddots& \vdots \\
S^{m1}_{|\mathcal{D}_m|1}  & \cdots &S^{m1}_{|\mathcal{D}_m||\mathcal{D}_1|}  &  &S^{mm}_{|\mathcal{D}_m|1}  & \cdots &S^{mm}_{|\mathcal{D}_m||\mathcal{D}_m|}
\end{bmatrix}
\in \mathbb{R}^{\theta\times \theta}.
\end{aligned}
\end{equation}
The objective function of (\ref{equ:derivative3}) can be rewritten by matrix form as follows,

\begin{equation}
\label{equ:derivative4}
\begin{aligned}
g(Z)
&= \left\|\bfeta-\bfv^\top Z\right\|^2_2+
2\lambda_2 Tr\left(Z\left(diag(\bf1^\top S) - S\right)Z^\top\right)\\
&+Tr\left(A^\top\left(Z-\overline{Z}\right)\right)
+\frac{\beta}{2}\left\|Z - \overline{Z}\right\|_2^2\\
&= \left[ \bfeta^\top \bfeta - 2 Tr(Z^\top \bfv \bfeta) + Tr(Z^\top \bfv \bfv^\top Z)\right]\\
&+2\lambda_2 Tr\left(Z\left(diag(\bf1^\top S) - S\right)Z^\top\right)
+Tr\left(\left(Z-\overline{Z}\right)^\top A\right)
\\
&
+\frac{\beta}{2}\left[Tr(Z^\top Z) - 2Tr(Z^\top \overline{Z}) + Tr(\overline{Z}^\top \overline{Z})\right ]\\
&= Tr(Z^\top \left(\bfv \bfv^\top+I\right) Z)+2\lambda_2 Tr\left(Z\left(diag(\bf1^\top S) - S\right)Z^\top\right)\\
&-  Tr\left(Z^\top \left(2\bfv \bfeta - A + \beta\overline{Z}\right)\right)+constant,
\end{aligned}
\end{equation}
where $\bf1\in R^\mathbb{\theta}$ is a vector of ones of $\theta$ dimensions, $diag(\bfx)$ is a diagonal matrix with the diagonal elements as the elements of vector $\bfx$, and $constant$ is a constant irrelevant to $Z$. To minimize the Lagrange function with regard to $Z$, we use the method of gradient descent. The matrix $Z$ is descended to the direction of gradient. The gradient function of $g(Z)$ is calculated as follows,

\begin{equation}
\label{equ:derivative5}
\begin{aligned}
&\nabla g(Z)
= 2\left(\bfv \bfv^\top+I\right) Z +4\lambda_2 Z\left(diag(\bf1^\top S) - S\right)-  \left(2\bfv \bfeta - A + \beta\overline{Z}\right).
\end{aligned}
\end{equation}
The updating rule is as follows,

\begin{equation}
\label{equ:update}
\begin{aligned}
Z^{new}\leftarrow Z^{old} - \psi &\nabla g(Z^{old}),
\end{aligned}
\end{equation}
where $\psi$ is the descent step size, and $\psi=\frac{1}{t}$ for the $t$-th iteration. The updating process is repeated until convergence.

\subsubsection{Update step for $\mathcal{W}_j$}

The third sub-optimization problem is the minimization of the Lagrange function with regard to the filters of CNN models. To update the filters, we only consider the terms of the Lagrange function relevant to the filters, and the following reduced Lagrange function is obtained,

\begin{equation}
\label{equ:LagrangianW}
\begin{aligned}
&J(\mathcal{W}_j)=
\lambda_1 \left (\sum_{j=1}^m \sum_{k=1}^{u} \|\bfw_k^j\|_2^2 \right )
\\
&+\sum_{j=1}^m\sum_{i=1}^{|\mathcal{D}_j|}\sum_{k=1}^{u}\alpha_{ik}^j\left([\bfz_i^j]_k - \max_{\tau=1}^{|\mathcal{X}_i^j|-h+1}\sigma({\bfw_k^j}^\top[\bfy_i^j]_\tau)\right)\\
&+\frac{\beta}{2}\sum_{j=1}^m\sum_{i=1}^{|\mathcal{D}_j|}\sum_{k=1}^{u}\left([\bfz_i^j]_k - \max_{\tau=1}^{|\mathcal{X}_i^j|-h+1}\sigma({\bfw_k^j}^\top[\bfy_i^j]_\tau)\right)^2.
\end{aligned}
\end{equation}
Since the obtained function is a combination of functions of independent filters, we propose to update the filter sequentially and independently. When one filter is considered, the others are fixed. The individual objective containing one single filter $\bfw_k^j$ is given as follows,

\begin{equation}
\label{equ:LagrangianW1}
\begin{aligned}
&J(\bfw_k^j)=
\lambda_1 \|\bfw_k^j\|_2^2
+\sum_{i=1}^{|\mathcal{D}_j|}\alpha_{ik}^j\left([\bfz_i^j]_k - \max_{\tau=1}^{|\mathcal{X}_i^j|-h+1}\sigma({\bfw_k^j}^\top[\bfy_i^j]_\tau)\right)\\
&+\frac{\beta}{2}\sum_{i=1}^{|\mathcal{D}_j|}\left([\bfz_i^j]_k - \max_{\tau=1}^{|\mathcal{X}_i^j|-h+1}\sigma({\bfw_k^j}^\top[\bfy_i^j]_\tau)\right)^2.
\end{aligned}
\end{equation}
Directly optimization of this problem is difficult, because it is composed of a maximization problem to seek the output of the CNN model. Instead of optimizing it directly, we develop an iterative algorithm to solve the problem alternately. We define an indicator to indicate which window gives the maximum response of $\sigma({\bfw_k^j}^\top[\bfy_i^j]_\tau)$ among $\tau=1,\cdots,|\mathcal{X}_i^j|-h+1$,

\begin{equation}
\label{equ:indicate}
\begin{aligned}
\tau_{ik}^j = {\arg\max}_{\tau=1}^{|\mathcal{X}_i^j|-h+1} \sigma({\bfw_k^j}^\top[\bfy_i^j]_\tau).
\end{aligned}
\end{equation}
In each iteration, we first update the indicators using the previous filter, and then fix indicators to update the filter. When the indicators are fixed, the problem in (\ref{equ:LagrangianW1}) is transformed to

\begin{equation}
\label{equ:LagrangianW2}
\begin{aligned}
&J(\bfw_k^j)=
\lambda_1 \|\bfw_k^j\|_2^2
+\sum_{i=1}^{|\mathcal{D}_j|}\alpha_{ik}^j\left([\bfz_i^j]_k - \sigma\left({\bfw_k^j}^\top[\bfy_i^j]_{\tau_{ik}^j}\right)\right)\\
&+\frac{\beta}{2}\sum_{i=1}^{|\mathcal{D}_j|}\left([\bfz_i^j]_k - \sigma\left({\bfw_k^j}^\top[\bfy_i^j]_{\tau_{ik}^j}\right)\right)^2.
\end{aligned}
\end{equation}
To minimize this objective, we also use the gradient descent method, and the gradient function is given as follows,

\begin{equation}
\label{equ:gradient1}
\begin{aligned}
&\nabla J(\bfw_k^j)=
2\lambda_1 \bfw_k^j
-\sum_{i=1}^{|\mathcal{D}_j|}\alpha_{ik}^j \nabla\sigma\left({\bfw_k^j}^\top[\bfy_i^j]_{\tau_{ik}^j}\right)[\bfy_i^j]_{\tau_{ik}^j}\\
&-\beta\sum_{i=1}^{|\mathcal{D}_j|}\left([\bfz_i^j]_k - \sigma\left({\bfw_k^j}^\top[\bfy_i^j]_{\tau_{ik}^j}\right)\right)
\nabla\sigma\left({\bfw_k^j}^\top[\bfy_i^j]_{\tau_{ik}^j}\right)[\bfy_i^j]_{\tau_{ik}^j}.
\end{aligned}
\end{equation}
where $\nabla\sigma(\cdot)$ is the gradient function of the nonlinear function $\sigma(\cdot)$. The updating rule for $\bfw^j_k$ is as follows,

\begin{equation}
\label{equ:updating}
\begin{aligned}
{\bfw_k^j}^{new} = {\bfw_k^j}^{old} - \psi \nabla J({\bfw_k^j}^{old}),
\end{aligned}
\end{equation}
where $\psi$ is the descent step size, and $\psi = \frac{1}{t}$ for the $t$-th iteration.

\section{Experiments}
\label{sec:exp}

In this section, we evaluate the proposed cross-model CNN method (CMCNN) experimentally. It is compare to some other cross-model representation methods.

\subsection{Benchmark data sets}

In the experiments, we use two multiple modality data sets to evaluate the proposed method. The data sets are described as follows.

\textbf{Wiki}. The first data set we used is the Wiki data set of two modalities, which are image and text. There are in total 2,866 documents in this data set. For each document, an image and a text is contained. To represent each image, we split the image to small patches and use a window to extract features from neighboring patches. The visual features of SIFT are extracted from each patch. The text is represented by the embeddings of the words. The number of classes of this data set is 10. A text/image is considered to be relevant to another text/image if they are in the same class.

\textbf{NUS-WIDE}. The second data set we use is the NUS-WIDE data of two modalities, which are also image and text. In this data set, there are 72,219 documenters of image-text pairs, belonging to 21 different classes. Similarly, the images are also presented by sequences of image patches, and the text is treated as sequence of words.


\subsection{Experimental setting}

To conduct the experiment of retrieval, we use the ten-fold cross-validation. Each data set is split to ten folds. Each fold is used as a query set, while the rest nine folds are used as training set and database set. The proposed method is applied to the training set to train the parameters, and then the parameters are used to represent both the data points of training and query set to CNN representations. The CNN representations of queries and database data points are compared by $\ell_2$ norm distance to rank the database data points. The top ranked data points are returned as the retrieval results.

The retrieval performance is measured by the performance of Precision at top $k$ (Prec@$k$), mean Average Precision (mAP), and the Break Even Point of Recall-Precision curve (BEPRP).The definition of these performance measures are given as follows.

\begin{equation}
\begin{aligned}
&Prec@k = \frac{\#\{database~ items~ ranked~ at~ top~ k~ \& ~relevant ~ to~ the~ query\}}{k},\\
&Reca@k = \frac{\#\{database~ items~ ranked~ at~ top~ k~ \& ~relevant~ to~ the query\}}{\#\{database~ items~ relevant ~ to~ the query\}},\\
&mAP = \frac{1}{Q}\sum_{q=1}^Q \frac{1}{D} \sum_{k=1}^D \left( Prec@k(q)\times\delta@k(q) \right),~and\\
&BEPRP = Prec@{k^*},~with~Prec@{k^*}=Reca@{k^*},k^*\in\{1,\cdots,D\},
\end{aligned}
\end{equation}
where Reca@$k$ is the Recall at top $k$, $Q$ is the total number of queries, $D$ is the number of database items. $Prec@k(q)$ is the Prec@$k$ of the $q$-th query, and $\delta@k(q)=1$ if the $k$-th database item is relevant to the $q$-th query, and $0$ otherwise.

\subsection{Experimental results}

\begin{figure}
  \includegraphics[width=0.3\textwidth]{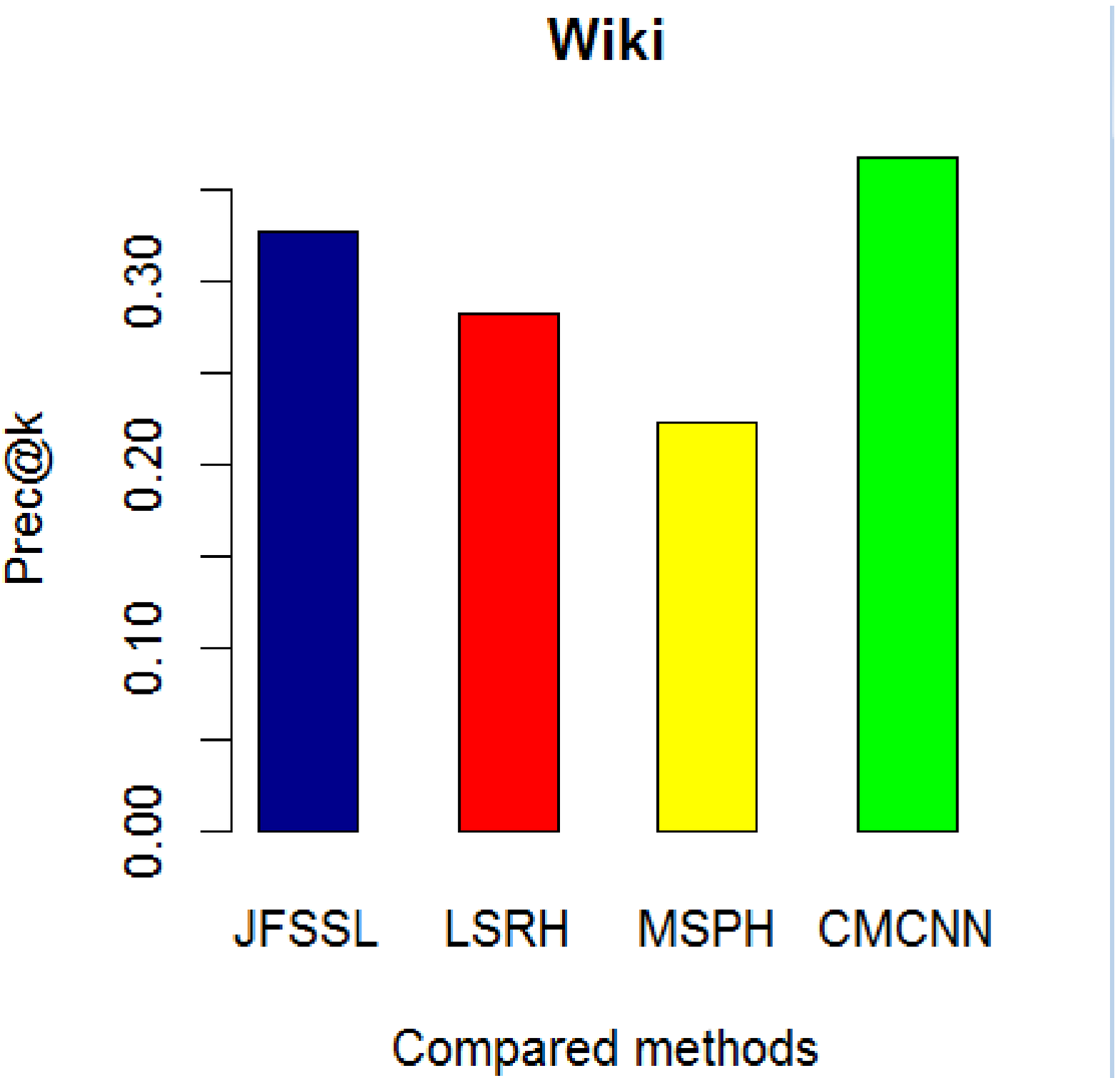}
  \includegraphics[width=0.3\textwidth]{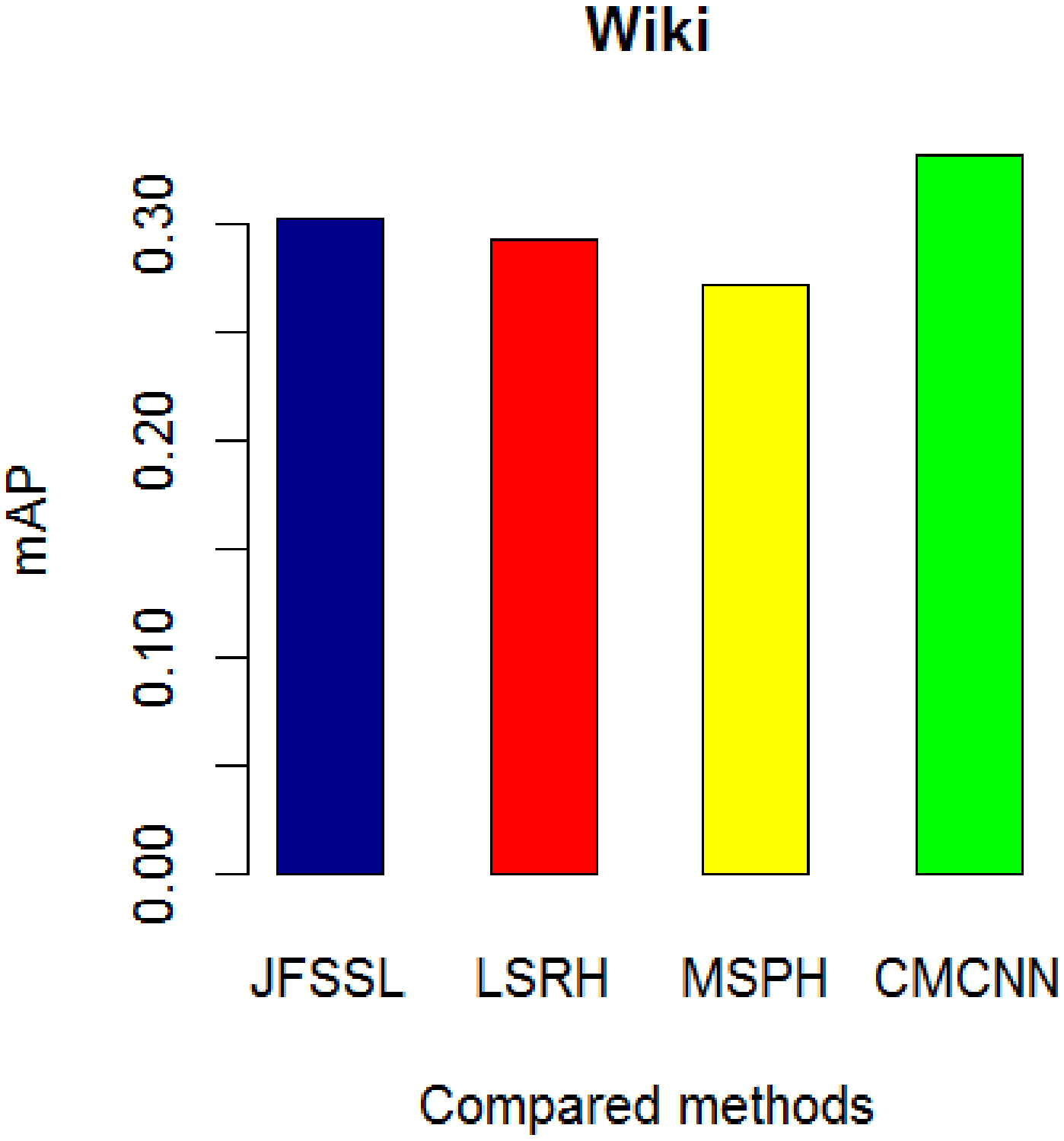}
  \includegraphics[width=0.3\textwidth]{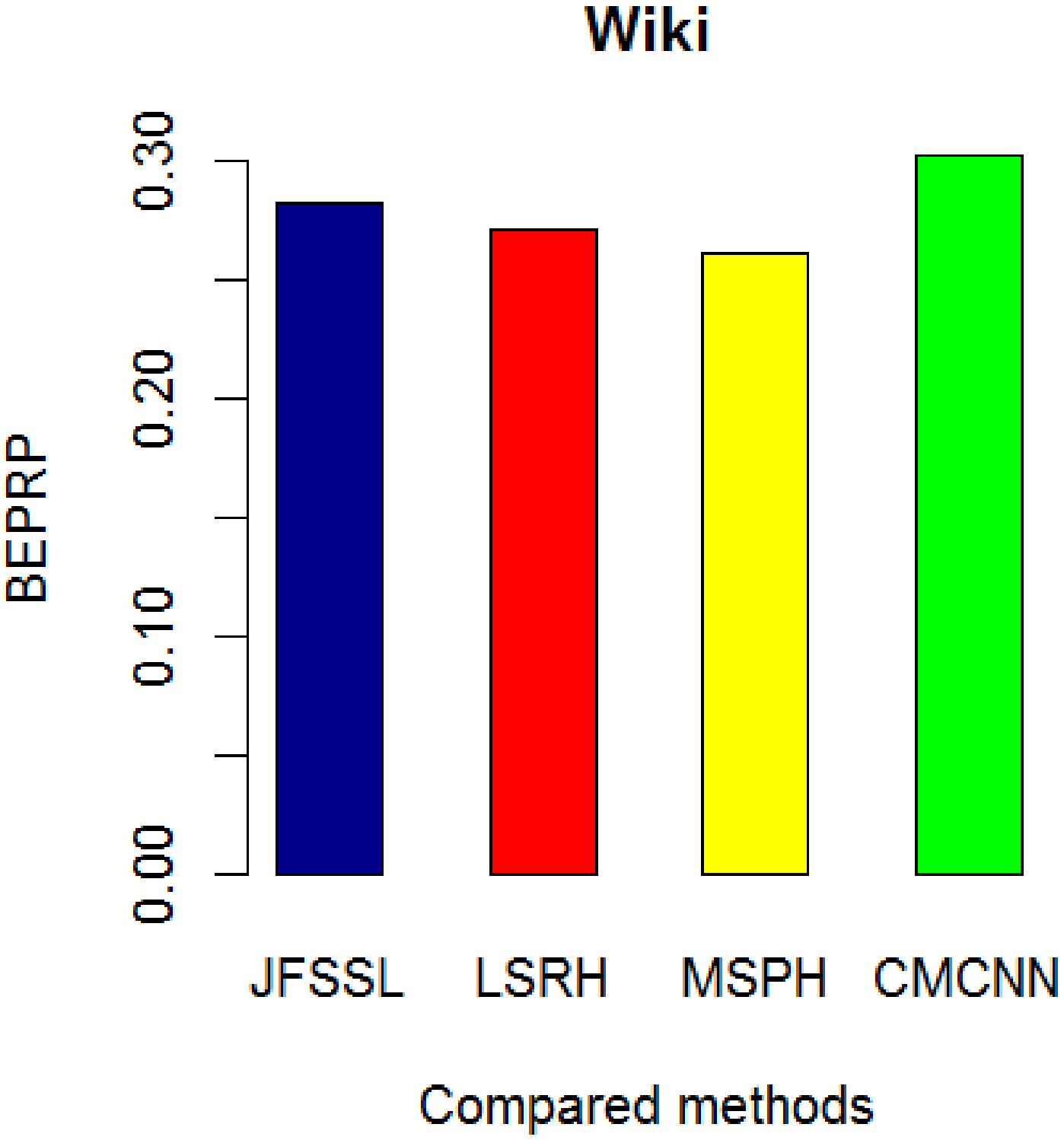}\\
  \includegraphics[width=0.3\textwidth]{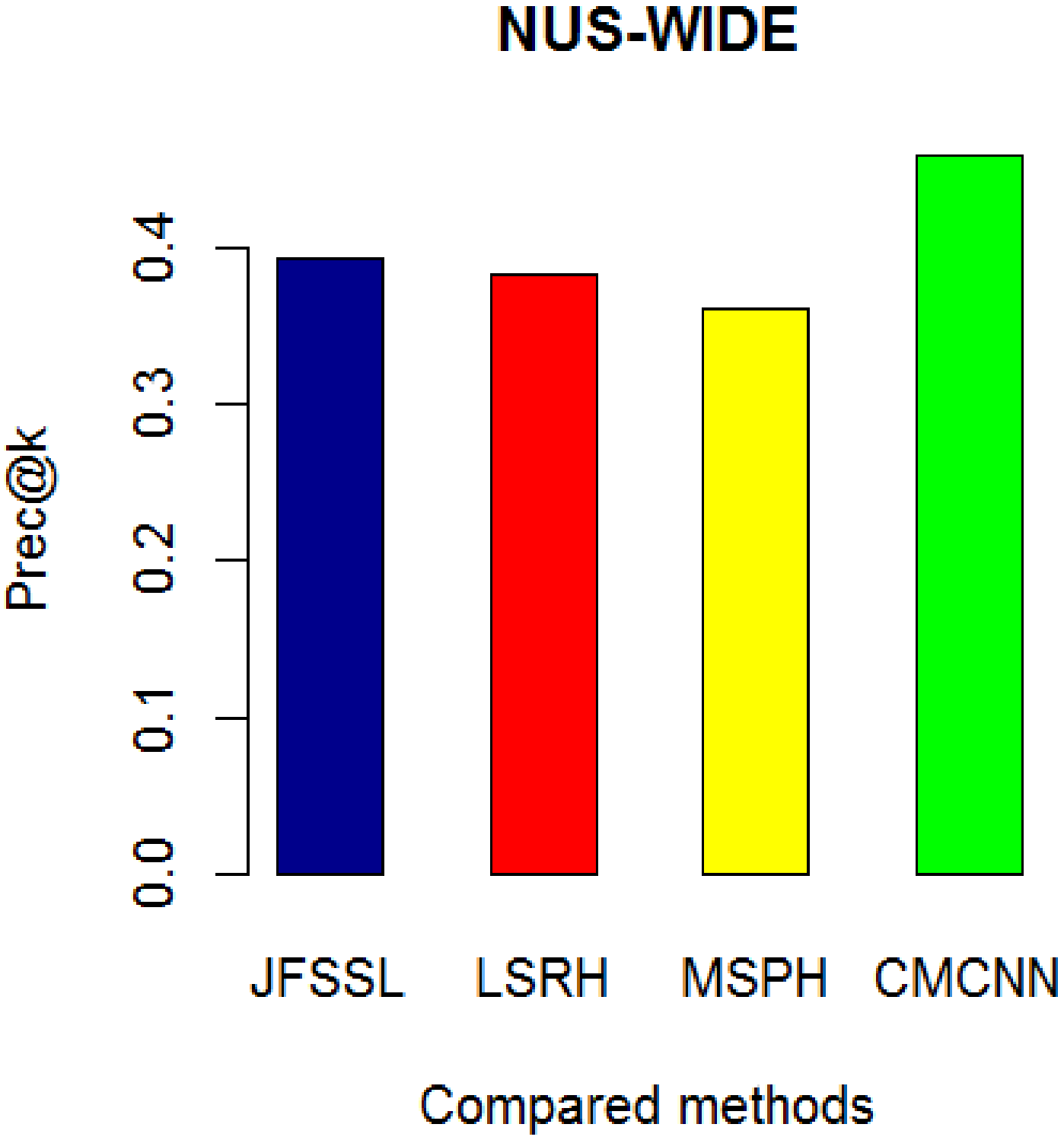}
  \includegraphics[width=0.3\textwidth]{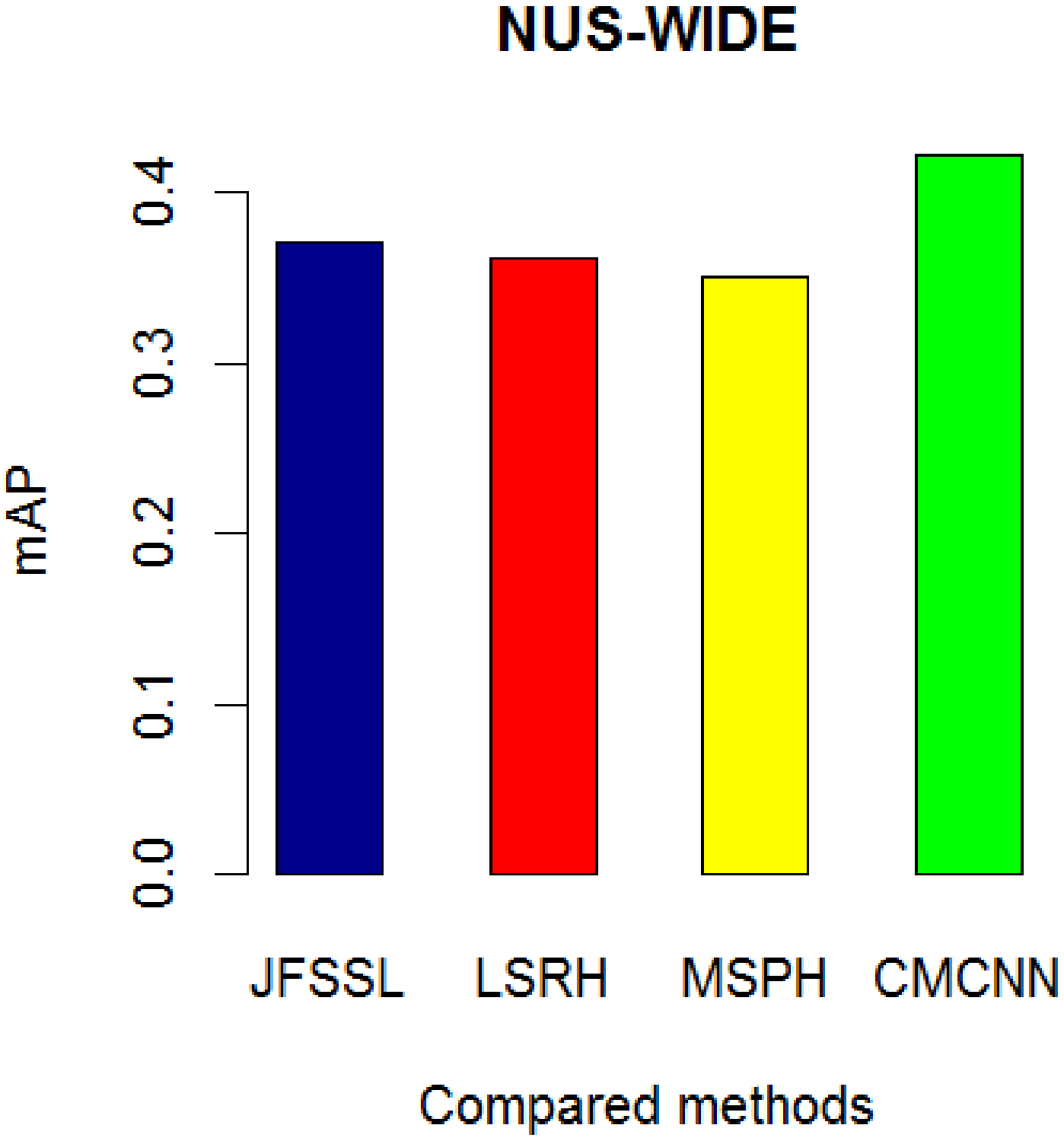}
  \includegraphics[width=0.3\textwidth]{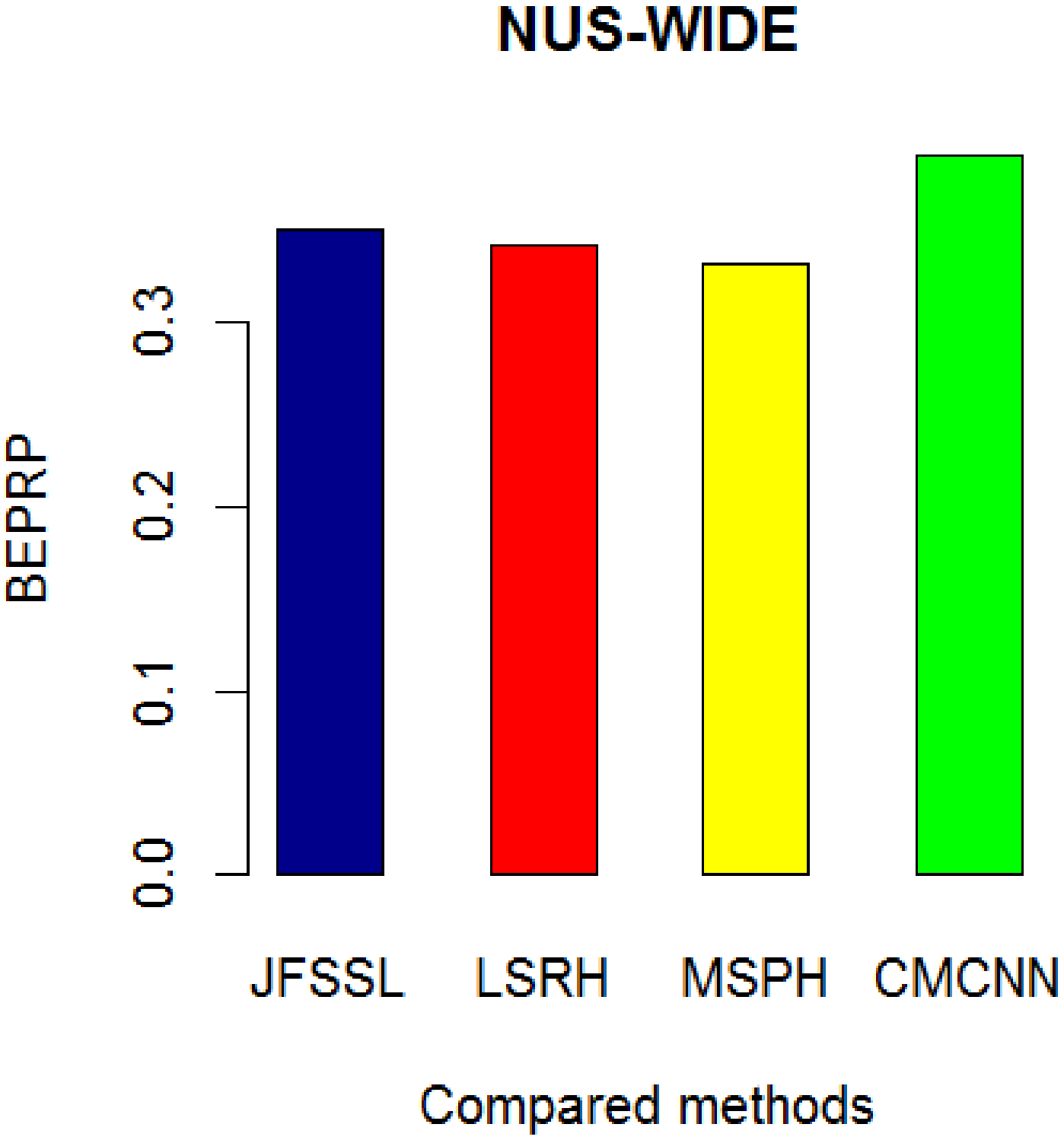}
  \caption{Comparison results over the benchmark data sets.}
  \label{fig:wiki}
\end{figure}

We compare the proposed method to the stat-of-the-art cross-model representation methods, including the cross-model Joint Feature Selection and Subspace Learning (JFSSL) \cite{He7346492}, the cross-modal Linear Subspace Ranking Hashing (LSRH) \cite{Li7571151}, and the Multimodal Similarity-Preserving Hashing (MSPH) \cite{masci2014multimodal}. The comparison results are reported in Figure \ref{fig:wiki}. According to the results reported in the figure, it is clearly that the proposed method CMCNN outperforms the compared cross-model representation methods significantly over all the three benchmark data sets. It is also noted that the outperforming of the method is also significant even measured by different performance measures. This is a further evidence of the advantage of the proposed method. Please note that because the compare methods, JFSSL, LSRH, and MSPH cannot take sequence data as input directly, thus we use vector quantization (VQ) method to transfer the sequences to histogram vectors, which are used as inputs of these methods. Thus the performance of the compared methods heavily relies on the quality of the histograms. However, the proposed CMCNN is a convolutional method which takes sequence data as input naturally. It has been shown that CNN is a good model to represent sequence data. This is a possible reason for the good performance of CMCNN.

\section{Conclusions}
\label{sec:conc}

In this paper, we propose to learn convolutional models to solve the problem of cross-model data representation and retrieval problems. We propose to map the data of different modalities to a common space, and in this commons space, classification and retrieval can be performed. Also in the common space, data of different modalities can be compared, and we use a cross-model relevance regularization to compare the dissimilarities of the data points. To map the sequences of different modalities, we propose to use the CNN model with multiple filters and max-pooling operations. Different modalities have different filter banks but can map the sequences to a common space. A linear classifier is applied to the outputs of the cross-model CNN to predict the class labels, and we also regularize the outputs of the CNN by a cross-model relevance term. The experiments over the benchmark data sets show its advantage over the existing cross-model data representation. In the future, we will also consider using the proposed method to other applications, such as integrated circuit design \cite{zhang2013eot,zhang2009new}, software engineering \cite{huang2011autoodc,geng2015improving,geng2014process}, network measurement \cite{chen2017data,chen2015differential,chen2016dispersing}, commuter vision \cite{zhu2011sparse,zhu2014scalable,chen2013computing,wang2014effective,liu2015supervised}, medical imaging \cite{li2015outlier,mo2015importance,king2015surgical,thatcher2016multispectral,li2015burn,squiers2016multispectral,dimaio2016reflective},  etc. We will also consider to use some other loss function to learn the parameters of the CNN and the classifier to optimize the multivariate performance measures \cite{wang2015multiple,liang2016optimizing,li2016nuclear,lin2016multi}.

\section*{Acknowledgements}

This work was supported by the Natural Science Foundation of Hebei Province (D2015207008), Talent Training Project of Hebei Province (A201400215) and Young Prominent Talent Project of Hebei Province Higher School (BJ2014021).

\section*{Statement of conflicts of interests}

The authors of this manuscript state that there is no conflicts of interests between this manuscript and other published works.


\end{document}